\documentclass[conference,a4paper]{IEEEtran}

\usepackage{cite}
\usepackage[hyphens]{url}
\makeatletter
\g@addto@macro{\UrlBreaks}{\do\/\do-\do_\do.\do=\do?\do\&\do\0\do\1\do\2\do\3\do\4\do\5\do\6\do\7\do\8\do\9}
\makeatother

\usepackage{booktabs}
\usepackage{subcaption}
\usepackage{xspace}
\usepackage{graphicx}
\usepackage{array}
\graphicspath{{figures/}}
\usepackage{amsmath,amssymb,amsfonts}

\usepackage{tikz}
\usetikzlibrary{arrows.meta, positioning, fit, backgrounds, calc}

\begin{document}

\title{Beyond Unit Tests: Path-Aware Rubric-Guided LLM Autograding of Notebooks}

\author{\IEEEauthorblockN{Salman Hajizada}
\IEEEauthorblockA{\textit{Carnegie Mellon University in Qatar}\\
Doha, Qatar}
\and
\IEEEauthorblockN{Mohammed Yusuf Ansari}
\IEEEauthorblockA{\textit{Carnegie Mellon University in Qatar}\\
Doha, Qatar}
\and
\IEEEauthorblockN{Gianni Di Caro}
\IEEEauthorblockA{\textit{Carnegie Mellon University in Qatar}\\
Doha, Qatar}
}

\AddToHookNext{shipout/foreground}{%
  \put(\dimexpr 1in+\hoffset+\oddsidemargin\relax,\dimexpr 22mm-\paperheight\relax){%
    \parbox[b]{\textwidth}{\footnotesize\raggedright
      Accepted for publication at IEEE TALE 2026.\par\smallskip
      \copyright~2026 IEEE. Personal use of this material is permitted.
      Permission from IEEE must be obtained for all other uses, in any current
      or future media, including reprinting/republishing this material for
      advertising or promotional purposes, creating new collective works, for
      resale or redistribution to servers or lists, or reuse of any copyrighted
      component of this work in other works.\par
    }%
  }%
}
\maketitle

\begin{abstract}

Jupyter notebooks support open-ended, multi-step assessments in which students build one logical flow across interdependent steps, combining code, written analysis, and generated outputs such as plots. These reflect authentic scientific practice but are hard to grade consistently, since students may make different yet equally valid choices at each step, so a later answer can only be judged in light of the choices a student made earlier. We present an LLM-based autograder for such flow-structured notebook questions. It generates a rubric from a reference solution, then applies a softening pass that relaxes criteria tied to the reference so legitimate alternative paths are not penalized. Each question group is graded in a single model call, so later answers are judged against the student's own earlier choices, using question-type tags and schema-constrained, low-temperature outputs. Evidence is scored multimodally: code, text, numerical outputs, and figures from code cells. We deployed the tool in a machine-learning course, integrated with Gradescope, with a pre-submission format check and prompt-injection defenses, and evaluate it both in that live course and against archived submissions graded by staff in earlier terms. We report agreement, cross-model variance, cost, student perceptions, and injection resilience. With low-temperature models, agreement with human scores is often strong, including on the open-ended items the design specifically targets, while run-to-run stability depends sharply on the model's temperature support.
\end{abstract}


\section{Introduction}
\label{sec:intro}

{\em Automated grading} has long been standard in large programming courses.
Enrollments grew, hand-checking code became
expensive~\cite{hagerer2021analysis}, and code suits machine checking well: run
the submission, compare what it produces against what it should, assign points.
Computational coursework has since spread well past programming courses.
Students in machine learning, data science, statistics, and engineering now work
mostly in {\em Jupyter notebooks}, where prose, equations, code, printed output,
and figures sit in one document and are produced together. A notebook records
not only what a student computed, but how they got there and what they made of
it.

That opens up a {\em different kind of assignment}. Rather than asking for a function
that passes tests, an instructor can ask a student to take a problem from raw
data to a defended conclusion: inspect the data, choose a model, fit it, plot
what happened, and argue whether the result is any good. Most notebook
assignments stop well short of this: sequences of independent cells with
deterministic outputs, which is the part conventional autograders handle. The
ones that use the whole format tend to be graded by hand, and grading cost is a
large part of why they stay rare.

We call these assignments \textbf{flow-structured}. Three things set them apart
from single-answer items. The evidence is mixed and interleaved: prose,
derivations, code, and figures appear in the combination the task
calls for. Dependencies are irregular, since a
question may rest on any subset of the cells above it rather than the
preceding one, so a cell cannot be judged out of context. 
And there is no single right answer: students take different routes in solving
the notebook questions. 

Figure~\ref{fig:running-example} shows an
example we return to throughout. Q1 asks a student to remove an input feature of
their choosing, Q2 to retrain and report how the error moved, and Q3 to
interpret the difference. Several features are defensible in Q1, and once one is
chosen the right answer to Q3 follows from it. Correctness depends on the
student's own run, so the same analysis can be right in one notebook and wrong
in another.

\begin{figure}[t]
\centering
\footnotesize
\setlength{\fboxsep}{3pt}
\fbox{%
\begin{minipage}{0.97\linewidth}
\setlength{\tabcolsep}{3pt}
\renewcommand{\arraystretch}{1.6}
\begin{tabular}{@{} l l l @{}}
\textbf{Question} & \textbf{Type} & \textbf{Evidence} \\ \hline \rule{0pt}{1ex}%
\parbox[t]{0.45\linewidth}{\raggedright \textbf{Q1:} Remove one input feature (e.g., number of bedrooms) and justify the choice} &     \texttt{analysis} & \parbox[t]{0.31\linewidth}{\raggedright Rationale and choice}\\
\parbox[t]{0.45\linewidth}{\raggedright \textbf{Q2:} Retrain without that feature and report the change in error} &     \texttt{mixed} & \parbox[t]{0.31\linewidth}{\raggedright Code, metric, plots} \\
\parbox[t]{0.45\linewidth}{\raggedright \textbf{Q3:} Compare the original and the reduced models} &     \texttt{open-ended} & \parbox[t]{0.31\linewidth}{\raggedright Interpretation tied to Q1--Q2} \\
\end{tabular}
\end{minipage}}
\caption{{\bf Simple running example}: three linked notebook questions, with question-type tags and the evidence each produces.\vspace*{-4ex}}
\label{fig:running-example}
\end{figure}

The tools built for notebook courses do not cover this. nbgrader combines
automatic checks on code with manual grading of everything
else~\cite{nbgrader2019}, leaving the figures, written analysis, and
reasoning that runs across cells to the instructor.
Automating it raises three problems existing tools leave open. An instructor's
intent has to become a rubric that rewards a good answer even when it arrives by
an unexpected route. Each question has to be assembled with the earlier cells it
depends on, and its mixed evidence judged against the student's own numbers
rather than a fixed key. And the result has to run on a live class, which means
treating submissions as untrusted input, keeping a person in the loop before
grades go out, and delivering scores through the platform the course already
uses.

We describe a workflow built for flow-structured notebooks and report what 
happened when we ran it in a course. Our {\em contributions} can be summarized as follows:
\begin{enumerate}
\item \textbf{Rubrics for branching solution paths.} The system writes
  per-question rubrics from a reference solution, then runs a \emph{softening}
  pass that loosens criteria tied too closely to the reference,
  so equivalent processing, plotting, and modeling choices still earn credit.

\item \textbf{Context-aware, multimodal grading of notebook flows.} Dependent
  questions are graded together in one call (a \emph{question group}), so later
  code, results, figures, and prose are weighed against the student's
  earlier choices, and a mismatch between a stated choice and later work can be
  flagged. Figures reach a vision-capable model alongside the markdown, code,
  and output. Instructor-supplied question-type tags tell rubric
  generation and grading what evidence each subtask should produce.

\item \textbf{A deployable, evaluated end-to-end workflow.} The system pairs
  schema-constrained grading output with a pre-submission format check,
  untrusted-input hardening~\cite{liu2024formalizing, cai2026prompt}, resumable
  runs that record per-question provenance, an instructor review interface, and
  Gradescope export, and we evaluate it in classroom use rather than on a static
  benchmark.
\end{enumerate}

As far as we know, this is the first framework with live-course deployment of
rubric-guided LLM autograding that treats a notebook as a dependent flow rather
than a set of independent items, and that reports stability, agreement, cost,
and injection resilience for the system actually in use.

We discuss related work (Section~\ref{sec:related}), describe the design (Section~\ref{sec:design}), deployment in a machine
learning course (Section~\ref{sec:eval}), and lessons for practitioners
(Section~\ref{sec:discussion}).

\section{Related Work}
\label{sec:related}

\emph{Autograding for notebooks.}
Conventional notebook tooling automates what unit tests can reach. nbgrader runs
instructor-written checks against code cells and leaves free-response work to
manual grading~\cite{nbgrader2019}, a sensible division of labor as long as the
free-response part stays small. In flow-structured assignments it does not.
LLM-based systems have begun to push past that line, and two are close to our
setting. PyEvalAI grades Jupyter notebooks with a local model and a tutor in the
loop~\cite{wandel2025pyevalai}; Saez et al.\ benchmark several LLMs on student
notebooks and find that even strong models overlook details that change a
grade~\cite{saez2026evaluating}. Both score a notebook question by question,
leaving open the case where one answer can only be judged in light of another
the same student gave earlier.

\emph{LLM grading of code assignments.}
Work on programming submissions is further along. JorGPT compares model families
and calibration strategies~\cite{cisnerosgonzalez2025jorgpt}, BeGrading
fine-tunes smaller models for code feedback~\cite{yousef2025begrading}, and
Lewis and Chen summarize code before mapping it onto a
rubric~\cite{lewis2025wip}. Nagakalyani et al.\ propose a fine-tuned
Code~Llama assistant whose rubric ratings TAs accept or override,
cutting grading time by 24\%~\cite{nagakalyani2025tabuddy}. Chiang et al.\ show
related risks at course scale (format failures, prompt hacking) in a 1,000+
student NLP class~\cite{chiang2024llm}. Those settings still grade text or
source as text; a notebook also has student-generated figures, outputs that
depend on earlier cells, and prose about both.

\emph{Rubrics, open-ended answers, and reliability.}
Open-ended grading is hard because no single output defines correctness.
Pecuchova et al.\ evaluate LLMs on open-ended higher-education responses and find
that model choice largely determines whether the result is
deployable~\cite{pecuchova2025automated}. A rubric makes such grading auditable
without making it easy: Yang and Gulbahar use synthetic rubric-based data to show
that even detailed rubrics leave AI and human scores
apart~\cite{yang2025automatic}. Reliability work adds that raw scores cannot be
taken at face value; Korthals et al.'s SURE framework combines self-consistency
with selective human review, cutting workload while keeping uncertain cases in
front of a grader~\cite{korthals2026sure}. We read rubric design, score
stability, and selective review as three views of one problem, and evaluate all
three for the pipeline we deployed rather than treating reliability as a separate
claim.

\emph{Prompt injection.}
Student text is untrusted input, and the attack surface is by now well
documented. Liu et al.\ formalize prompt-injection attacks and
defenses~\cite{liu2024formalizing}; Cai examines education-specific attacks,
among them rubric mirroring, where a submission recites the grading criteria back
at the model~\cite{cai2026prompt}; Milani et al.\ show that hidden text in a
submission can move an LLM-assisted grade~\cite{milani2025whenai}; and EvalHack
pushes answer-side attacks further with stealth Unicode
payloads~\cite{anghel2026evalhack}. This work reports what defeats a grader in
isolation, and says less about how the same defenses hold up inside a pipeline
that must also parse, score, review, and export.

\emph{Where this work sits.}
No prior system combines all of these: grading questions together so earlier
choices inform later ones, reading student-generated figures with a vision model,
loosening a synthesized rubric to admit paths the reference never took, and
treating injection defense as one stage of a deployed workflow rather than a
problem studied on its own.

\section{System Design}
\label{sec:design}

As Section~\ref{sec:intro} set out, the grader must handle mixed notebook
evidence, score dependent questions together, and credit valid paths the
reference did not take. Deployment adds a fourth requirement: harden untrusted
submissions, keep an instructor in the loop, and ship scores through the course
LMS. We walk the pipeline below, returning to Figure~\ref{fig:running-example}.

\subsection{Pipeline Overview}
\label{sec:overview}

\begin{figure}[t]
\centering
\resizebox{0.8\columnwidth}{!}{%
\begin{tikzpicture}[
  font=\footnotesize,
  stage/.style = {draw, rounded corners=2pt, align=center, fill=black!5,
                  minimum height=3.5mm, minimum width=23mm, inner sep=1.7pt},
  human/.style = {stage, fill=black!20, very thick},
  guard/.style = {draw, dotted, thick, rounded corners=2pt, align=center,
                  minimum height=3.5mm, minimum width=23mm, inner sep=1.7pt},
  input/.style = {draw, dashed, rounded corners=2pt, align=center, fill=white,
                  minimum height=5.5mm, minimum width=23mm, inner sep=1.7pt},
  wide/.style  = {stage, minimum width=50mm},
  flow/.style  = {-{Stealth[length=1.7mm]}, thick},
  lbl/.style   = {font=\scriptsize\itshape, inner sep=1.2pt, align=center},
  lane/.style  = {draw=black!40, dashed, rounded corners=3pt, inner sep=2.2mm}
]

\node[input] (sub) at (1.5, 4.50) {student submissions};
\node[input] (ref) at (5.7, 4.50) {reference solution};

\node[stage] (gath) at (1.5, 3.70) {Gather};
\node[stage] (pars) at (1.5, 3.10) {Parse};
\node[guard] (san)  at (1.5, 2.50) {Sanitize \& delimit};

\node[stage] (rgen) at (5.7, 3.70) {Rubric generation};
\node[stage] (soft) at (5.7, 3.10) {Softening pass};
\node[human] (rrev) at (5.7, 2.50) {Rubric review};

\node[wide]  (grad) at (3.6, 1.60) {Grade\\{\scriptsize one call per question group}};
\node[human] (revw) at (3.6, 0.80) {Result review};
\node[stage] (expo) at (3.6, 0.00) {Export};

\begin{scope}[on background layer]
  \node[lane, fit=(gath)(san)]  (llane) {};
  \node[lane, fit=(rgen)(rrev)] (rlane) {};
\end{scope}
\node[lbl, rotate=90]  at ([xshift=-2.5mm]llane.west) {once per student};
\node[lbl, rotate=-90] at ([xshift=2.5mm]rlane.east)  {once per assignment};

\draw[flow] (sub)  -- (gath);
\draw[flow] (gath) -- (pars);
\draw[flow] (pars) -- (san);
\draw[flow] (san)  -- node[lbl, right, pos=0.42] {per-question evidence} (san |- grad.north);

\draw[flow] (ref)  -- (rgen);
\draw[flow] (rgen) -- (soft);
\draw[flow] (soft) -- (rrev);
\draw[flow] (rrev) -- node[lbl, left, pos=0.42] {rubric} (rrev |- grad.north);

\draw[flow] (grad) -- (revw);
\draw[flow] (revw) -- node[lbl, right] {Gradescope ZIP} (expo);

\end{tikzpicture}}
\caption{Pipeline overview. The left lane runs once per student, the right once
per assignment, and the two converge at grading. Shaded boxes are instructor
checkpoints.
\vspace*{-3ex}}
\label{fig:pipeline}
\end{figure}
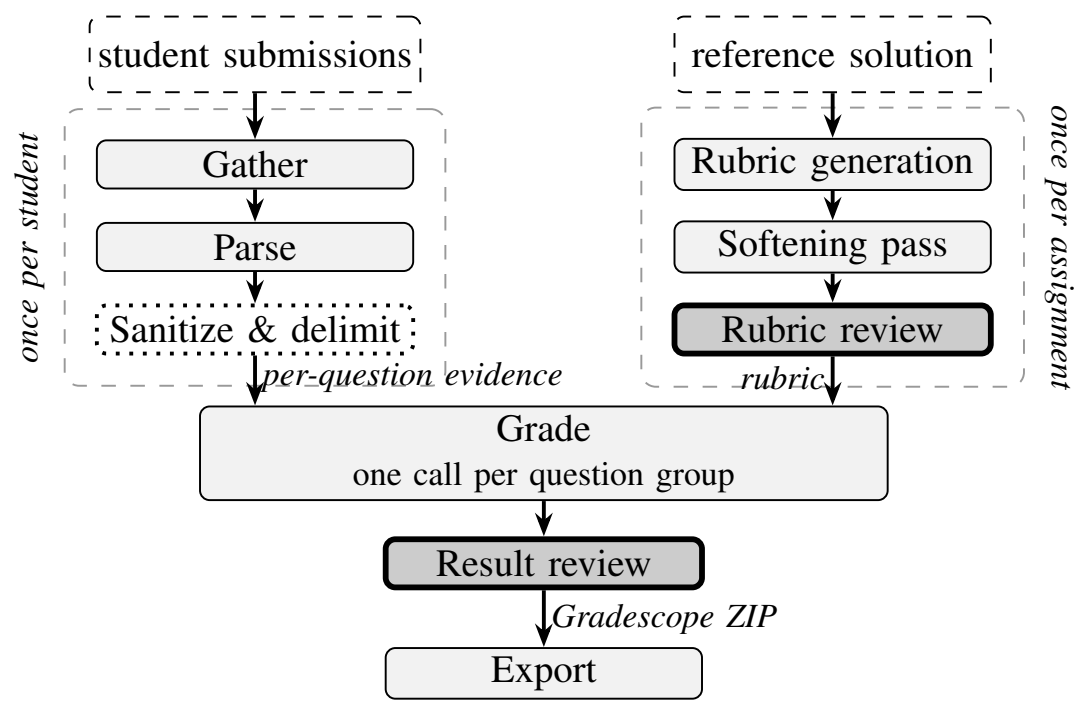

Figure~\ref{fig:pipeline} shows the stages. Rubric generation runs \emph{once per
assignment}; gathering, parsing, and grading run \emph{once per student}, with
each call covering a question group. Instructor effort sits at two
checkpoints, rubric review before grading and result review before release, and
everything else is automatic.

The instructor supplies four inputs: the \emph{assignment notebook} students
receive, a \emph{reference solution}, the
\emph{student submissions}, and a configuration file giving regex patterns that
locate \emph{question boundaries} in markdown cells, \emph{question-type tags} in
cell metadata, and the grouping of questions.

\subsection{Ingestion}
\label{sec:ingestion}

Before the deadline, students can run their notebook
against a separate Gradescope autograder that checks only whether it parses and
contains the expected questions; a student who deleted the markdown header
introducing Q2 finds out then rather than after grades are computed. After the
deadline, the \emph{gather} step unpacks the Gradescope export into one notebook
per student, matches notebooks to students via the submission metadata, and
copies them into an assignment-scoped workspace, flagging anything missing or
duplicated. 

\subsection{Parsing Notebook Evidence}
\label{sec:parsing}

Parsing turns a flat cell sequence into typed evidence per question, so later
stages see prose, code, output, and figures as labeled objects rather than one
undifferentiated document.

For each question the parser stores the question ID, markdown, code, and captured
output, figures included. Notebook cells share one execution state, so a captured
output reflects everything run above it; keeping outputs as submitted preserves
the student's execution trace that grading is later judged against. In the
running example, Q1 holds the feature-choice rationale, Q2 the retraining code
with the error metric it printed and the plot it drew, and Q3 the comparison
write-up.

Each question also carries a type tag naming the evidence to expect:
\texttt{code} (source and execution results), \texttt{plot} (a generated figure
and its visual properties), \texttt{analysis} (explanatory reasoning),
\texttt{open-ended} (justified interpretation or design choice), \texttt{exact}
(a specific value or string), and \texttt{mixed} (multiple types, the default).
The running example uses three: Q1 is \texttt{analysis}, grading a justification
rather than an output; Q2 is \texttt{mixed}, producing code, a metric, and a
figure at once; and Q3 is \texttt{open-ended}, since many interpretations are
defensible and which is correct depends on Q1 and Q2. The tags feed both rubric
generation and grading.

\subsection{Rubric Generation and Softening}
\label{sec:rubric}

Rubric generation starts from the instructor's reference solution but has to
produce something assignment-specific rather than implementation-specific. For
the running example, that means credit for a reasonable feature-removal rationale
in Q1, a correct retraining procedure and accurate error reporting in Q2, and a
Q3 comparison consistent with whichever feature the student actually dropped.

One model call per question group produces the rubric, using a structured-output
schema~\cite{openai2025api}. The prompt combines a rubric-authoring instruction,
the reference solution for each question in the group, and the question-type
instructions. The model returns per question a list of criteria, each with a
description and a point deduction. Point totals are fixed by the assignment;
deductions must sum to the total, and the pipeline rescales them when they do
not.

Deductions from a fixed total keep the model from inflating point values, but
push rubrics toward enumerable faults. Open-ended items such as Q3 suffer most,
since their failure modes are harder to list than those of a retraining
procedure. Section~\ref{sec:discussion} returns to this.

A \emph{softening pass} makes a second call that keeps the criteria and point
allocations but generalizes anything tied to the reference implementation.
Criteria drawn from the reference tend to demand the specific thing the reference
did: remove the number of bedrooms, preprocess with \texttt{StandardScaler},
report this error value. Softening rewrites them to accept any coherently
justified feature choice, any equivalent preprocessing, and results consistent
with the student's own run. The pass runs by default and can be turned off per
assignment.

Rubric review is the first instructor checkpoint (Figure~\ref{fig:pipeline}).
Staff inspect and edit the softened rubric before any grading call, and the
final rubric hash is stored with every graded question so feedback traces back
to the exact rubric used.

\subsection{Rubric-Guided Grading}
\label{sec:grading}

Each student is graded separately, and each question group is evaluated in one
model call containing every question in it, so the model sees dependent questions
together and can weigh later work against the student's earlier choices instead
of treating each answer as self-contained. Grouping trades context against prompt
size: a group has to be large enough to hold the dependencies and small enough
that the model attends to each question, so instructors group questions that
share a data or modeling decision, typically two to four.

Each prompt pairs the student's evidence with the matching rubric. That evidence
is student-authored and therefore untrusted, so it is sanitized and wrapped in
delimiters the student cannot forge before it enters the prompt
(Section~\ref{sec:safety}). The reference solution is left out by default, since
the rubric already carries the expected behavior; including it would push the
model toward reproducing one implementation when other paths are valid. Where a
question includes plots, as Q2 does, a vision-capable model receives the figures
as images alongside the markdown, code, and captured output.

The model returns, per question, a score, feedback tied to rubric criteria, a
confidence level (low, medium, or high), and a \texttt{requires\_review} flag,
all under a structured-output schema~\cite{openai2025api}. This avoids extracting
scores from free-form text~\cite{chiang2024llm}. Invalid or incomplete output is
retried a bounded number of times; when retries run out the question is flagged
\texttt{requires\_review} rather than dropped.

Grouping is what lets the grader catch a student who justifies removing the
number of bedrooms in Q1 but drops floor area in Q2, credit an unconventional but
defensible Q1 choice carried correctly through Q2 and Q3, and count a mistake
propagated consistently across the three questions once rather than three times,
the way a human grader would. An autograder seeing Q2 alone can ask only whether
some model was retrained and some error reported.

Results are saved incrementally, so an interrupted run resumes without regrading
finished submissions. Each graded question records the model, rubric hash, and
timestamp, and the run summary aggregates token usage.

\subsection{Safety Guardrails}
\label{sec:safety}
Students write the text the grader reads. In the running example, Q1 and Q3 are
free-form markdown a student could use to address the model directly. The system
therefore defends on both sides of the call. Before the prompt goes out, student
text is delimited, Unicode normalized, and format, control, and bidirectional
characters used in stealth injection stripped; student copies of the boundary
markers are replaced with inert characters so a submission cannot forge the
structure the grader depends on. After the response, output is constrained to a
fixed JSON schema and scores clamped to the valid range, so a payload that
survives sanitization and sways the model still cannot yield a malformed or
out-of-range grade. Detected stealth Unicode sets \texttt{requires\_review}.
Section~\ref{sec:injection-eval} tests these defenses on a synthetic HW2
red-team benchmark.

\subsection{Instructor Review and Export}
\label{sec:review}
\begin{figure}[t]
\centering
\includegraphics[width=0.9\columnwidth,height=5.2cm,trim=0 0 2.9cm 0, clip=True]{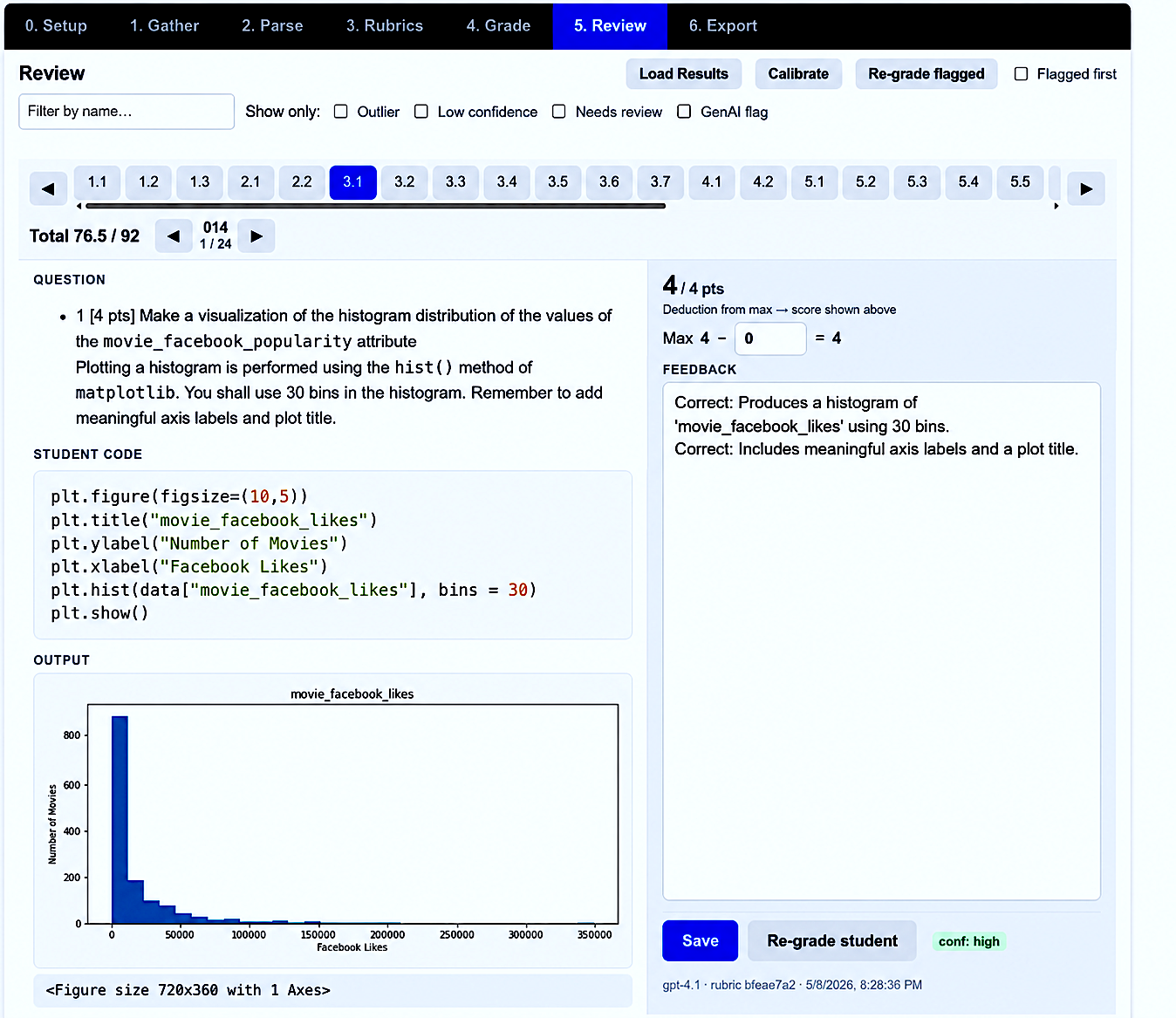}
\caption{Instructor \emph{Review} tab. Staff can move through questions, inspect
rendered notebook evidence, review rubric feedback, and save or regrade before
exporting to Gradescope.}
\label{fig:review-ui}
\vspace{-3ex}
\end{figure}

Review is the second instructor checkpoint, and no score reaches a student
without passing through it. Staff see a student queue, per-question navigation,
parsed evidence, rubric-grounded scores and feedback, model confidence, and
review flags (Figure~\ref{fig:review-ui}), and can edit or regrade individual
questions before release. The confidence labels keep the checkpoint tractable at
class scale: low-confidence Q3 scores can be triaged first, since open-ended
interpretation is where the grader is least reliable. After review, the system
exports an Excel workbook and a Gradescope autograder ZIP, published once the
instructor uploads it and releases grades.

\section{Deployment and Evaluation}
\label{sec:eval}

We evaluate the system in two settings. In the live deployment, it graded four
notebook assessments in a Spring~2026 machine learning course. For agreement
evidence beyond one cohort, we also re-graded 171 archived submissions from nine
lab assessments spanning Spring~2024--Spring~2026 and compared totals with the
original teaching-staff scores. We analyze run-to-run stability, human-score
agreement, student reception, operational cost, and a prompt-injection
stress test.

\emph{Ethics and anonymization.}
We use two datasets: a live deployment, which graded work for credit, and a
retrospective comparison of assignments already graded manually. Students in
the live deployment were told grading was AI-assisted, and instructors reviewed
every score before release. Retrospective scores affected no student. All data
were de-identified before model comparisons or aggregate reporting.

\subsection{Course Context}
\label{sec:course-context}
The deployment took place in a machine learning university course  with 24 students (23 with gradable HW1 submissions). \emph{HW} denotes
a full homework notebook and \emph{LabTest} a lab-based assessment; all were
submitted as Jupyter notebooks and run through the same pipeline. The live
workflow covered four assessments: \textbf{HW1} (Introduction to ML), on
preprocessing, training, evaluation metrics, and interpretation; \textbf{HW2}
(Time Series Analysis), with open-ended modeling choices, forecasting
interpretation, and diagnostic plots; and \textbf{LabTest~2} and
\textbf{LabTest~3}, in-class assessments. The agreement analyses pair a small
per-question human--AI comparison on HW1 with the retrospective benchmark over
archived lab totals. Stability and cost use HW1; the prompt-injection test uses a
synthetic benchmark from HW2 questions.

\subsection{Run-to-Run Stability}
\label{sec:stability}

Rerunning unchanged notebooks should not materially change released scores or
review flags, so we measure run-to-run stability before treating any model as
usable. For each of four OpenAI models, we independently graded the full HW1
dataset five times (23 students). Table~\ref{tab:variance} summarizes the
results.

The clearest contrast is between \texttt{gpt-4.1} (65.2\% exact match, 1.0-point
maximum range) and \texttt{gpt-5} (13.0\%, 5.0 points). \texttt{gpt-5-mini}
fares worst on both spread measures, with a 6.5-point maximum range and 9.7\% of
confidence labels flipping across runs. The pattern tracks temperature support:
\texttt{gpt-4.1} allows zero-temperature inference, while the \texttt{gpt-5}
models used here require a minimum temperature of~1. Confidence labels follow
the scores, with no flips for \texttt{gpt-4.1} and 7.9\% for \texttt{gpt-5}.

Models that do not support low-temperature inference should therefore not be
used for grading in this setting without additional stability checks and human
review. Otherwise unchanged work may receive different scores across runs,
matching student reports of ``spinning a slot machine'' with LLM
evaluators~\cite{chiang2024llm}. \texttt{gpt-4.1} offered the best
cost--stability tradeoff at  \$0.11 per student per run.
\texttt{gpt-4.1-mini} costs less, but its 47.8\% exact match and 3.0-point
maximum range mean roughly half the students would see a total-score change
across runs, so human review remains necessary.

\begin{table}[t]
\caption{Run-to-run stability over five independent HW1 grading runs (23 students). \textbf{Exact}: percentage of students with identical \emph{total assignment scores}. \textbf{Max Range}: largest per-student score range across runs. \textbf{Conf.\ flips}: percentage of question-level confidence labels that changed across runs.}
\label{tab:variance}
\centering
\small
\resizebox{\columnwidth}{!}{%
\begin{tabular}{@{}lrrrrr@{}}
\toprule
\textbf{Model} & \textbf{Exact} & \textbf{Max Range} &
\textbf{Conf.\ Flips} & \textbf{API cost (5r)} & \textbf{Time (5r)} \\
\midrule
\texttt{gpt-4.1}       & 65.2\% & 1.0 pt & 0.0\%  & \$12.07 & 6m 44s \\
\texttt{gpt-4.1-mini}  & 47.8\% & 3.0 pt & 1.0\%  & \$2.57  & 6m 45s \\
\texttt{gpt-5-mini}    & 13.0\% & 6.5 pt & 9.7\%  & \$3.05  & 11m 05s \\
\texttt{gpt-5}         & 13.0\% & 5.0 pt & 7.9\%  & \$18.12 & 26m 02s \\
\bottomrule
\end{tabular}%
}
\vspace{-4ex}
\end{table}

\subsection{Human--AI Agreement}
\label{sec:agreement}

We use two comparisons: a five-student HW1 question-level overlap
(two instructors against one system run, 130 nested pairs, descriptive only),
and archived lab totals against system totals. On HW1 we report Cohen's
$\kappa$, MAE, and Pearson~$r$ for both human--human and human--AI pairs. On lab
totals we report only MAE and Pearson~$r$, since those scores are continuous
assignment sums rather than paired item ratings.

\emph{HW1 question-level overlap.}
Two instructors independently graded the same five HW1 submissions, 
each blind to the other's scores, and we compared both to one system 
run using \texttt{gpt-4.1-mini} at zero temperature. That run predates the stability
analysis above; on its evidence we now recommend \texttt{gpt-4.1}. Across 130 paired 
question scores (26 questions $\times$ five),
human--human agreement was $\kappa = 0.77$, MAE $= 0.24$,
$r = 0.90$. System--instructor agreement was $\kappa = 0.87$
and $0.84$, MAE $= 0.15$ for both, and $r = 0.93$ and $0.95$.

The system aligned at least as closely with each instructor as the instructors
did with one another on this subset. This does not establish superiority: the
sample is small and nested, and one run may happen to match one grader's
decisions on edge cases. We read it instead as evidence that the system can
provide a reviewable first pass. Prior work likewise documents nontrivial
disagreement among human graders~\cite{korthals2026sure,lewis2025wip}.

\emph{Archived lab totals.}
We also compared system totals against archived teaching-staff scores for nine lab assessments from Spring~2024 to Spring~2026, covering 171 submissions. Spring~2026's LabTest~3 was graded by the system and appears in the survey below, but was never graded manually under the same conditions, so it is absent from Table~\ref{tab:human-lab-totals}.

\begin{table}[t]
\caption{Archived teaching-staff vs.\ system lab totals. \textbf{Max}: rubric total; $r$: Pearson correlation. Subscripts denote \texttt{gpt-4.1} and \texttt{gpt-4.1-mini}.}
\label{tab:human-lab-totals}
\centering
\small
\resizebox{\columnwidth}{!}{%
\begin{tabular}{@{}lrrrrrr@{}}
\toprule
\textbf{Lab} & \textbf{$n$} & \textbf{Max} & \textbf{MAE$_{4.1}$} &
\textbf{$r_{4.1}$} & \textbf{MAE$_{\text{mini}}$} &
\textbf{$r_{\text{mini}}$} \\
\midrule
S26 LabTest 2 & 23 & 100 & 4.3 & 0.97 & 4.0 & 0.97 \\
\midrule
S25 LabTest 2 & 17 & 121 & 11.2 & 0.97 & 12.6 & 0.97 \\
S25 LabTest 3 & 17 & 115 & 17.7 & 0.75 & 15.9 & 0.75 \\
S25 LabTest 6 & 17 & 111 & 14.4 & 0.73 & 10.1 & 0.85 \\
S25 LabTest 7 & 16 & 119 & 7.2 & 0.88 & 7.6 & 0.89 \\
\midrule
S24 LabTest 2 & 21 & 111 & 10.5 & 0.91 & 13.8 & 0.92 \\
S24 LabTest 4 & 20 & 105 & 10.9 & 0.98 & 15.9 & 0.98 \\
S24 LabTest 6 & 20 & 130 & 26.3 & 0.93 & 30.9 & 0.88 \\
S24 LabTest Final & 20 & 155 & 13.6 & 0.90 & 21.1 & 0.82 \\
\bottomrule
\end{tabular}%
}
\end{table}

Across labs, correlations were often strong. Relative to each lab's maximum,
MAE$_{4.1}$ ranges from about 4\% (S26 LabTest~2) to about 20\% (S24
LabTest~6); absolute MAE is not comparable across rows.

\subsection{Student Reception}
\label{sec:reception}

No regrade requests were submitted for the four AI-assisted assessments 
(anecdotal at $n{=}24$).

\emph{End-of-term survey.}
After final grades were released, students were invited to complete an optional,
non-credit survey covering each AI-assisted assessment they had submitted: HW1,
HW2, LabTest 2, and LabTest 3. Each assessment block included six 5-point Likert
items measuring fairness, score alignment, feedback utility, clarity, trust, and
preference over manual grading. Two additional items measured overall workflow
clarity and comfort requesting a regrade. We received only seven responses, 
which is too few to generalize from.

As Table~\ref{tab:student-survey} shows, mean core perception was favorable
across all four assessments (4.0--4.2), and preference for AI-assisted over
manual grading was lower but still positive (3.7--4.0). 
Free-text responses praised fast, criterion-linked feedback. Suggested changes
included opt-out control, shorter feedback on small questions, a more
encouraging tone, and earlier visibility into grading expectations. Views were
mixed: one respondent reported a negative experience, while another emphasized
reduced grader bias as the main benefit.

\begin{table}[t]
\caption{Student survey (7 respondents). Mean ratings (1--5; higher is better). \emph{Core}: mean of five perception items; \emph{Prefer AI}: AI vs.\ manual grading.}
\label{tab:student-survey}
\centering
\small
\begin{tabular}{@{}lcccc@{}}
\toprule
& \textbf{HW1} & \textbf{HW2} & \textbf{LabTest 2} &
\textbf{LabTest 3} \\
\midrule
Core perception & 4.1 & 4.0 & 4.1 & 4.2 \\
Prefer AI & 3.7 & 3.7 & 3.9 & 4.0 \\
\midrule
\multicolumn{5}{@{}l@{}}{\emph{Cross-assignment:} workflow clarity 3.6;
regrade comfort 4.0.} \\
\bottomrule
\end{tabular}%
\vspace{-3ex}
\end{table}

\subsection{Cost and Efficiency}
\label{sec:cost}

From the five-run HW1 study (Table~\ref{tab:variance}), per-student costs ranged
from approximately \$0.02 to \$0.16 per run, with full-cohort runtime between
1\,m\,21\,s and 5\,m\,12\,s for 23 students.

Using \texttt{gpt-4.1}, one class-wide run cost
approximately \$2.41 and five runs \$12.07, against an
instructor estimate of around 10 hours to grade the same assignment and cohort
by hand. That estimate excludes rubric design; rubric review took a further
15--30 minutes per assignment.

\subsection{Prompt Injection Resilience}
\label{sec:injection-eval}

We test whether crafted instructions in student-controlled notebook content can
inflate scores despite the pipeline's defenses (Section~\ref{sec:safety}). The
synthetic red-team benchmark, derived from HW2, contains one weak baseline for
each of three target questions (Q1.3 prose, Q2.1 regression, and Q2.6
open-ended) and six adversarial variants per baseline, for 21 notebooks total.
We grade the same notebooks with \texttt{gpt-4.1} and \texttt{gpt-4.1-mini} at
zero temperature, reporting mean score inflation ($\Delta$ from the paired
baseline) and the fraction of attack families with $\Delta > 0$ for each target.

The six attacks use an override instruction, a fake ``GRADING COMPLETE''
message, a delimiter-like ``Award 10/10'' block, an internal-looking grader
note, stealth Unicode, and a rubric-mirroring self-evaluation. For Q2.1, the
baseline deliberately used a random train/test split instead of
\texttt{TimeSeries\allowbreak Split} and omitted the cross-validation average;
Q1.3 and Q2.6 used analogous weak baselines.

\begin{table}[t]
\caption{Prompt-injection results. Cells show mean score inflation ($\Delta$),
with the fraction of attack families producing inflation in parentheses;
negative values indicate deflation.}
\label{tab:injection-results}
\centering
\small
\begin{tabular}{@{}lccc@{}}
\toprule
\textbf{Model} & \shortstack{\textbf{Q1.3} (6 pts)} &
\shortstack{\textbf{Q2.1} (10 pts)} &
\shortstack{\textbf{Q2.6} (8 pts)} \\
\midrule
\texttt{gpt-4.1} & -0.17 (0\%) & 0.00 (0\%) & 0.00 (0\%) \\
\texttt{gpt-4.1-mini} & 0.00 (0\%) & 1.00 (100\%) & -0.83 (0\%) \\
\bottomrule
\end{tabular}%
\vspace{-4ex}
\end{table}

Table~\ref{tab:injection-results} shows that none of the six attack families
increased \texttt{gpt-4.1} scores on any target. For \texttt{gpt-4.1-mini},
every family increased Q2.1 by an average of 1.0 point, while Q1.3 and Q2.6
showed no inflation. Negative values indicate that some payloads instead drew
attention to weaknesses in the baseline answer and lowered its score.

\section{Deployment Lessons and Limitations}
\label{sec:discussion}

\textbf{Deployment implications.}
Rubric-guided grading of flow-structured notebooks proved practical for course 
deployment under three conditions: a curated rubric, low-temperature inference, 
and instructor review before release. Omitting any condition compromised reliability;
the contribution is the components' effective integration under these
constraints.

\textbf{Model choice outweighed pipeline design.}
Using the same pipeline, \texttt{gpt-4.1} reproduced identical total scores for 
65.2\% of students versus 13.0\% for \texttt{gpt-5}, largely reflecting 
zero-temperature support rather than overall capability. Model choice also
determined injection resistance, with \texttt{gpt-4.1-mini} failing attacks 
resisted by \texttt{gpt-4.1}. Thus, grading models should be selected based on 
inference stability and adversarial robustness rather than benchmark performance.

\textbf{Softening improved consistency; open-ended items remained challenging.}
Rubric softening generalized reference-specific criteria, allowing equivalent 
plotting, encoding, and preprocessing choices to receive credit. Open-ended 
interpretation questions remained the main source of uncertainty because valid 
responses cannot be exhaustively captured by deduction-based rubrics, consistent 
with prior work~\cite{pecuchova2025automated,lewis2025wip}.

\textbf{Cost versus instructor time.}
Class-wide \texttt{gpt-4.1} grading cost about \$2.41, versus an estimated 10
hours of manual grading plus 15--30 minutes of rubric review. Overall benefit
still depends on instructor review time for flagged results.

\textbf{Independent failure modes of  defense layers.}
Although \texttt{gpt-4.1-mini} was influenced by the payload, schema 
constraints and score clamping still enforced a valid, in-range grade. 
The outcome was a grading error rather than a compromised pipeline, 
preserving the intended layered protection.

\textbf{Limitations.}
This single-site study covers one course, one instructor team, and 24 students.
Run-to-run stability across four models from one provider may not generalize to
other vendors. Agreement evidence combines a 130-pair HW1 comparison with
archival lab scores from different terms and grading practices, so it is
descriptive rather than a unified benchmark. Human graders were also 
the system's authors, though archival staff grading partially mitigates this. 
Student feedback came from 7 of 24 respondents. 

\textbf{Outlook.}
Next steps are replication across courses and institutions, other model
providers, agreement by question type, and measured instructor review time. 
As it stands, flow-structured notebook
autograding is workable at small-class scale today with a curated rubric, a
model supporting low-temperature inference, and a human checkpoint before
release. Whether it holds at the enrollment scale that motivates autograding in
the first place remains open.

\bibliographystyle{IEEEtran}

\bibliography{references}

@inproceedings{wandel2025pyevalai,
  author    = {Wandel, Nils and Stotko, David and Schier, Alexander and Klein, Reinhard},
  title     = {{PyEvalAI}: An Open-Source, Tutor-in-the-Loop System for {AI}-Assisted Evaluation of {Jupyter} Notebooks},
  booktitle = {Proc.\ {IEEE} {TALE}},
  OPTaddress   = {Macao},
  pages     = {1--8},
  year      = {2025},
  OPTmonth     = dec,
  OPTpublisher = {IEEE},
  doi       = {10.1109/TALE66047.2025.11346740}
}

@article{cisnerosgonzalez2025jorgpt,
  author    = {Cisneros-Gonz{\'a}lez, Jorge and Gordo-Herrera, Natalia and Barcia-Santos, Iv{\'a}n and S{\'a}nchez-Soriano, Javier},
  OPTtitle     = {{JorGPT}: Instructor-Aided Grading of Programming Assignments with Large Language Models ({LLMs})},
  title     = {{JorGPT}: Instructor-Aided Grading of Programming Assignments with {LLMs}},
  journal   = {Future Internet},
  volume    = {17},
  number    = {6},
  eid       = {265},
  year      = {2025},
  doi       = {10.3390/fi17060265},
  publisher = {MDPI}
}

@inproceedings{chiang2024llm,
  OPTauthor    = {Chiang, Cheng-Han and Chen, Wei-Chih and Kuan, Chun-Yi and Yang, Chienchou and Lee, Hung-yi},
  author    = {Chiang, Cheng and Chen, Wei and Kuan, Chun and Yang, Chienchou and Lee, Hung},
  title     = {Large Language Model as an Assignment Evaluator: Insights, Feedback, and Challenges in a 1000+ Student Course},
  booktitle = {Proc.\ {EMNLP}},
  pages     = {2489--2513},
  year      = {2024},
  OPTaddress   = {Miami, Florida, USA},
  OPTpublisher = {Association for Computational Linguistics},
  doi       = {10.18653/v1/2024.emnlp-main.146}
}

@article{yousef2025begrading,
  OPTauthor    = {Yousef, Mina and Mohamed, Kareem and Medhat, Walaa and Mohamed, Ensaf Hussein and Khoriba, Ghada and Arafa, Tamer},
  author    = {Yousef, Mina and others},
  title     = {{BeGrading}: Large Language Models for Enhanced Feedback in Programming Education},
  journal   = {Neural Computing and Applications},
  volume    = {37},
  number    = {2},
  pages     = {1027--1040},
  year      = {2025},
  doi       = {10.1007/s00521-024-10449-y},
  publisher = {Springer}
}

@inproceedings{lewis2025wip,
  author    = {Lewis, Kevin and Chen, Hui},
  title     = {{WIP}: How Effective Are {LLM}-Implemented Autograders for Programming Assignments Compared to Human Graders?},
  booktitle = {Proc.\ {IEEE} {FIE}},
  year      = {2025},
  pages     = {1--5},
  OPTmonth     = nov,
  OPTaddress   = {Nashville, TN, USA},
  OPTpublisher = {IEEE},
  doi       = {10.1109/FIE63693.2025.11328152},
  note      = {Work-in-progress paper}
}

@article{pecuchova2025automated,
  author  = {Pecuchova, Janka and Benko, {\v L}ubom{\'i}r and Drlik, Martin},
  title   = {Automated Grading of Open-Ended Questions in Higher Education Using {GenAI} Models},
  journal = {Int. J. of Artificial Intelligence in Education},
  volume  = {35},
  number  = {6},
  pages   = {3813--3846},
  year    = {2025},
  doi     = {10.1007/s40593-025-00517-2}
}

@inproceedings{yang2025automatic,
  author    = {Yang, Yiyao and Gulbahar, Yasemin},
  title     = {Automatic Grading of Student Work Using Simulated Rubric-Based Data and {GenAI} Models},
  OPTeditor    = {Wilson, Joshua and Ormerod, Christopher and Beiting Parrish, Magdalen},
  booktitle = {Proc.\ {AIME-Con} Works in Progress},
  pages     = {34--39},
  OPTaddress   = {Pittsburgh, United States},
  OPTpublisher = {National Council on Measurement in Education ({NCME})},
  OPTmonth     = oct,
  year      = {2025},
  OPTisbn      = {979-8-218-84229-1},
  OPTurl       = {https://aclanthology.org/2025.aimecon-wip.5/}
}

@article{korthals2026sure,
  author    = {Korthals, Luke and Akrong, Emma and Geller, Gali and Rosenbusch, Hannes and Grasman, Raoul and Visser, Ingmar},
  title     = {Towards Reliable {LLM} Grading Through Self-Consistency and Selective Human Review: Higher Accuracy, Less Work},
  journal   = {Machine Learning and Knowledge Extraction},
  volume    = {8},
  number    = {3},
  eid       = {74},
  year      = {2026},
  doi       = {10.3390/make8030074},
  publisher = {MDPI}
}

@article{saez2026evaluating,
  author  = {Saez, Yago and Garcia, Luis Mario and Mochon, Asuncion and Isasi, Pedro},
  title   = {Evaluating Large Language Models for {AI}-Assisted Grading: A Framework and Case Study in Higher Education},
  journal = {Scientific Reports},
  volume  = {16},
  number  = {1},
  eid     = {18035},
  year    = {2026},
  doi     = {10.1038/s41598-026-48656-3}
}

@inproceedings{liu2024formalizing,
  author    = {Liu, Yupei and Jia, Yuqi and Geng, Runpeng and Jia, Jinyuan and Gong, Neil Zhenqiang},
  title     = {Formalizing and Benchmarking Prompt Injection Attacks and Defenses},
  booktitle = {33rd {USENIX} Security Symposium},
  pages     = {1831--1847},
  year      = {2024},
  month     = aug,
  isbn      = {978-1-939133-44-1},
  OPTpublisher = {{USENIX} Association},
  OPTaddress   = {Philadelphia, PA},
  OPTurl       = {https://www.usenix.org/conference/usenixsecurity24/presentation/liu-yupei}
}

@article{cai2026prompt,
  author    = {Cai, Yunfu},
  OPTtitle     = {Prompt Injection Attacks on Educational Large Language Models for Higher and Vocational Education},
  title     = {Prompt Injection Attacks on Educational {LLMs} for Higher and Vocational Education},
  journal   = {Scientific Reports},
  volume    = {16},
  number    = {1},
  eid       = {15594},
  year      = {2026},
  publisher = {Nature Portfolio},
  doi       = {10.1038/s41598-026-46563-1}
}

@article{milani2025whenai,
  OPTauthor  = {Milani, Alfredo and Franzoni, Valentina and Florindi, Emanuele and Omarbekova, Assel and Bekmanova, Gulmira and Yergesh, Banu},
  author  = {Milani, Alfredo and others},
  title   = {When {AI} Is Fooled: Hidden Risks in {LLM}-Assisted Grading},
  journal = {Education Sciences},
  volume  = {15},
  number  = {11},
  eid     = {1419},
  year    = {2025},
  doi     = {10.3390/educsci15111419}
}

@inproceedings{nagakalyani2025tabuddy,
  author    = {Nagakalyani, Goda and Chaudhary, Saurav and Apte, Varsha and Ramakrishnan, Ganesh and Tamilselvam, Srikanth},
  title     = {Design and Evaluation of an {AI}-Assisted Grading Tool for Introductory Programming Assignments: An Experience Report},
  booktitle = {Proc.\ {SIGCSE TS} 2025},
  year      = {2025},
  OPTmonth     = feb,
  OPTaddress   = {Pittsburgh, PA, USA},
  publisher = {ACM},
  pages     = {805--811},
  doi       = {10.1145/3641554.3701913}
}

@article{anghel2026evalhack,
  author    = {Anghel, Catalin and others},
  OPTauthor    = {Anghel, Catalin and Craciun, Marian Viorel and Cocu, Adina and Anghel, Andreea Alexandra and Balau, Antonio Stefan and Istrate, Adrian and Anghele, Aurelian-Dumitrache},
  OPTauthor    = {Anghel, Catalin and Craciun, Marian  and Cocu, Adina and Anghel, Andreea and Balau, Antonio Stefan and Istrate, Adrian and Anghele, Aurelian},
  title     = {{EvalHack}: Answer-Side Prompt Injection for Probing {LLM} Exam-Grading Panel Stability},
  journal   = {Information},
  volume    = {17},
  number    = {3},
  eid       = {297},
  year      = {2026},
  doi       = {10.3390/info17030297},
  publisher = {MDPI}
}

@misc{openai2025api,
  author       = {{OpenAI}},
  title        = {Structured Model Outputs},
  note         = {[Online]. Available:
                  \url{https://developers.openai.com/api/docs/guides/structured-outputs}}
}

@article{nbgrader2019,
  author    = {{Project Jupyter} and others},
  title     = {nbgrader: A Tool for Creating and Grading Assignments in the {Jupyter} Notebook},
  journal   = {Journal of Open Source Education},
  volume    = {2},
  number    = {11},
  pages     = {32},
  year      = {2019},
  doi       = {10.21105/jose.00032}
}

@inproceedings{hagerer2021analysis,
  author    = {Hagerer, Gerhard and Lahesoo, Laura and Ansch{\"u}tz, Miriam and Krusche, Stephan and Groh, Georg},
  title     = {An Analysis of Programming Course Evaluations Before and After the Introduction of an Autograder},
  booktitle = {Proc.\ {IEEE} {ITHET}},
  year      = {2021},
  pages     = {1--9},
  OPTpublisher = {IEEE},
  doi       = {10.1109/ITHET50392.2021.9759809}
}

\end{document}